# Improving Primary Healthcare Workflow Using Extreme Summarization of Scientific Literature Based on Generative AI

Gregor Stiglic
Faculty of Health Sciences
University of Maribor
Maribor, Slovenia
gregor.stiglic@um.si

Primoz Kocbek
Faculty of Health Sciences
University of Maribor
Maribor, Slovenia
primoz.kocbek@um.si

Pablo Meyer
Center for Computational Health,
IBM Thomas J. Watson Research
Center
Yorktown Heights
NY, USA
pmeyerr@us.ibm.com

Leon Kopitar
Faculty of Health Sciences
University of Maribor
Maribor, Slovenia
leon.kopitar1@um.si

Lucija Gosak
Faculty of Health Sciences
University of Maribor
Maribor, Slovenia
lucija.gosak2@um.si

Jiang Bian
Department of Health Outcomes
and Biomedical Informatics
College of Medicine, University of
Florida
FL, USA
bianjiang@ufl.edu

Zhe He
School of Information
Florida State University,
Tallahassee
FL, USA
zhe@fsu.edu

Prithwish Chakraborty
Amazon Science
New York
NY, USA
prithwich@gmail.com

## ABSTRACT

Primary care professionals struggle to keep up to date with the latest scientific literature critical in guiding evidence-based practice related to their daily work. To help solve the above-mentioned problem, we employed generative artificial intelligence techniques based on large-scale language models to summarize abstracts of scientific papers. Our objective is to investigate the potential of generative artificial intelligence in diminishing the cognitive load experienced by practitioners, thus exploring its ability to alleviate mental effort and burden. The study participants were provided with two use cases related to preventive care and behavior change, simulating a search for new scientific literature. The study included 113 university students from Slovenia and the United States randomized into three distinct study groups. The first group was assigned to the full abstracts. The second group was assigned to the short abstracts generated by AI. The third group had the option to select a full abstract in addition to the AI-generated short summary. Each use case study included ten retrieved abstracts. Our research demonstrates that the use of generative AI for literature review is efficient and effective. The time needed to answer questions related to the content of abstracts was significantly lower in groups two and three compared to the first group using full abstracts. The results, however, also show significantly lower accuracy in extracted knowledge in cases where full abstract was not available. Such a disruptive technology could significantly reduce the time required for healthcare professionals to keep up with the most recent scientific literature; nevertheless, further developments are needed to help them comprehend the knowledge accurately.

## 1 Introduction

Health system improvement in primary care is typically slow, partly due to not only the reliance on passive knowledge dissemination but also the absence of a systematic approach to identify the gaps between evidence and practice and to implement appropriate interventions to close these gaps [1,2]. Integrating innovations and evidence-based practices (EBPs) into routine care settings is the core of Learning Health Systems (LHSs). In an LHS, the EBPs should integrate the evidence synthesized from existing scientific literature with evidence generated from the data obtained from the institution's data networks [3]. This study leveraged generative AI approaches that use large language models (LLMs) for the summarization of scientific paper abstracts to allow fast screening of novel literature for an LHS in primary healthcare settings. Different scenarios were simulated to estimate the time savings using extreme summarization for literature screening.

DSHealth at SIGKDD '23, August 6 – 10, 2023, Long Beach, CAG. Stiglic et al.## 2 Methods

Participants in our study were provided with two use cases (UC1, UC2) related to preventive care and behavior change, and a search for relevant scientific literature was simulated (Figure 1, Figure 2). The case studies were developed within the Erasmus+ project "*Improving healthcare students' competences for behaviour change to effective support self-care in chronic diseases*"[1].

**Name:** Liam O'Malley
**Age:** 54 years old
**Life course:** Working age adult
**Need:** Chronic conditions
**Connectivity:** Broadband and mobile device
**Country:** Ireland
**Gender:** Male
**Job:** Plumber

Liam is a 54-year-old plumber who lives with his husband, Callum O'Reilly, in the outskirts of Dublin. Liam's husband is 52 years old and works at the local theatre as an assistant. Liam has always been very active both at work and in his social activities. He used to play football competitively when he was younger. He kept playing with his friends three times per week until three years ago, when he stopped due to a knee injury. Now he spends most of his energy at work; one of his hobbies are do-it-yourself projects in his garage workshop. Liam loves entertaining and chatting with friends around the table. Amidst friends, the couple is renowned for Callum's reinventions of Sunday roasts, with great meat cuts, fries and homemade sauces. Liam has gained weight after stopping playing football and was diagnosed with type 2 diabetes (T2D) and hypertension, two years ago.

**Unmet needs |** Liam is comfortable with his lifestyle. Albeit the difficulty in squatting worries him, he considers it isn't related to his health status. He has received lifestyle changing recommendations from his GP and diabetes nurse specialist, including diet and physical activity, but has not given it a serious thought. Frequently, he forgets to take his pills in the morning due to his busy workdays; sometimes he takes them in the evening and occasionally skips the medication. For Liam changing lifestyle behaviours is not a priority. However, his diabetes nurse recommended him a structured diabetes education program that caught his attention and perhaps could be a good opportunity to better manage his own health.

**Figure 1:** Use case 1 (UC1).

**Name:** Luuk de Vries
**Age:** 72 years old
**Life course:** Retired person
**Need:** Chronic conditions/complex needs
**Connectivity:** Broadband and mobile device
**Country:** Netherlands
**Gender:** Male

Luuk is a 72-year-old retired baker who lives with his wife, Marije, in a nice residential area of Utrecht. They have one son, Kjeld de Vries, and one grandson. Luuk's grandson is 23 years old and has just started his PhD in Lisbon University. Luuk was always active, both at work where he needed to be standing or moving around, and in his social activities. He also commuted daily to work on his bike. Since he retired his son runs the bakery business, but Luuk likes to oversee the quality control of the products to continue to have the best pastries in the region. After Luuk got married he started developing overweight. He was diagnosed with obesity ten years ago. His cholesterol and blood pressure have both increased since then and he started pharmacological therapy. Lifestyle changes have also been recommended by his healthcare professionals. Five years ago Luuk had a myocardial infarction, which damaged a considerable portion of the cardiac muscle and left him with heart failure. Luuk used to ride his single speed "oma's fiets" (Dutch bike) every day. Since he developed heart failure he does not feel capable of doing that anymore. Luuk gets fatigued very quickly. He has to sit during short walks and after taking a bath.

**Unmet needs |** Luuk is uncomfortable with his lifestyle. Luuk may need support in planning for the future such as hiring a homebased care service to help him and his wife manage their chronic conditions. A food provision service for Luuk and household help services would be helpful due to the progression of Luuk's chronic condition. Luuk also feels the need to learn more about, and to cope better with, his condition due to increasing fatigue. He also makes mistakes with medicines; such as duplicating doses and changing the times he takes his medicines.

**Figure 2:** Use case 2 (UC2).

Each use case resulted in 10 abstracts obtained from the Semantic Scholar platform along with the corresponding automatically generated extremely short summaries tagged with "TLDR" (short for Too Long, Didn't Read), which often consist of only one or two sentences in most cases [4]. Semantic scholar TLDRs are generated using the CATTS (Controlled Abstraction for TLDRs with Title Scaffolding) - a simple yet effective learning strategy for producing TLDRs that utilizes titles as a supplementary training signal. As shown in our previous research [5], general purpose LLMs such as OpenAI or Pegasus could also be used for such extreme summarization tasks resulting in similar performance. Each participant was asked to complete a set of 10 tasks (questions prepared by a group of primary healthcare experts) per use case and answer one question per abstract by choosing one of the four possible answers (Figure 3). The questions were asked according to the content presented in the abstract and could be answered by reading it.

ABSTRACT 8:

Alcohol is a previously unrecognized acute suppressor of serum Arg and lifestyle modification lowers asymmetric dimethylarginine in subjects who achieve weight loss >5%.

By at least how many percent should we reduce body weight to reduce ADMA?
- 1%
- 3%
- 5%
- more than 5%

Click checkbox if you need more detailed version of abstract: ☑

ABSTRACT 8:

Since low serum l-arginine (Arg) and high asymmetric dimethylarginine (ADMA) can predict microvascular complications in type 2 diabetes mellitus (T2DM), we tested whether Arg and ADMA are affected by diet and physical activity in overweight/obese and T2DM subjects. We tested the effects on serum Arg and ADMA of single loads of dextrose, protein, fat, or alcohol (~300 calories each); one episode of physical exercise; and 12 weeks of standard lifestyle modification (dietary and physical activity counseling). Alcohol drink was followed by ~30% lowering in Arg. Arg and ADMA increased after a protein load but remained stable after glucose or fat load or 30 min of treadmill walk. Following 12 weeks of lifestyle modification, ADMA declined only in subjects achieving weight loss >5%. In conclusion, alcohol is a previously unrecognized acute suppressor of serum Arg. Lifestyle modification lowers ADMA in subjects who achieve weight loss >5%.

Continue

**Figure 3: An example of a question for human evaluation.**

Higher education students with at least basic healthcare knowledge were enrolled in the study. Health sciences students were invited by one of the researchers to take part in the study. Students from all four years of study were invited to take part in the survey. The difficulty of the questions depended on the abstract used. Using a simple randomization procedure, they were divided into 3 study groups: (1) G1 - Full Abstract group that received full abstracts without summarization, (2) G2 - Short AI Generated Abstract with Optional Full Abstract group that received the short summaries and could expand the full abstract if needed, and (3) G3 - Short AI Generated Abstract group that received only the short summaries (Figure 4).

The questionnaire consisted of two parts. The first part contained demographic questions (gender, age, major, year of study) and the second part contained questions related to the content of the scientific abstracts.

---
[1] http://www.train4health.eu/

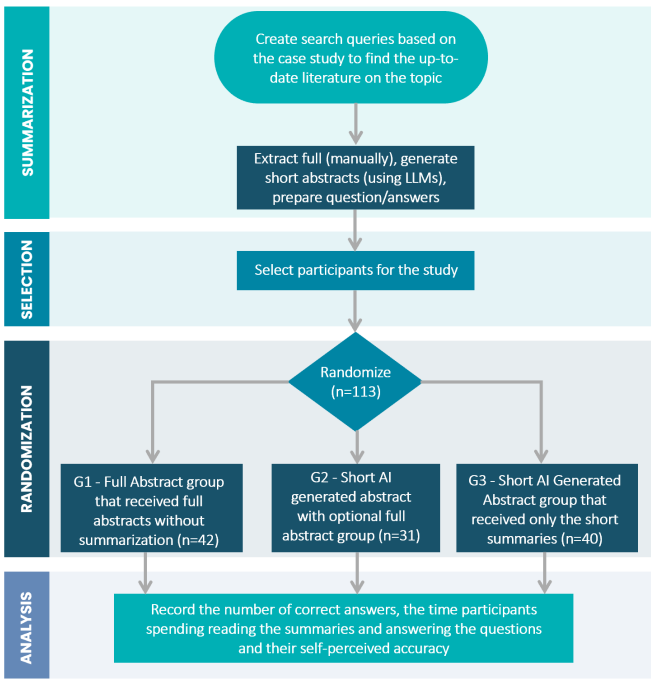

**Figure 4: Graphical overview of the study protocol.**

## 3 Results

Data from 113 participants (75 from the EU and 38 from the US) were collected. The survey was completed by 36 (31.9%) men and 77 (68.1%) women. 82 (72.6%) were enrolled in an undergraduate program, of whom 43 (38.1%) were in the second year and 39 (34.5%) in the third year. 77 (68.1%) students were enrolled in a postgraduate study program, of whom 21 (18.6%) were in the 1st year, 5 (4.4%) in the 2nd year, and 5 (4.4%) in the 3rd year. At the end of the questionnaire, students also answered how confident they were in answering the question correctly on a 10-point Likert scale. The mean value was 6.21 (SD=2.250).

In terms of accuracy in answering the questions, a significant difference between US and EU students can be observed, where the gap can be 20% and more in G2 and G3.

Since the sample sizes for EU and US students were imbalanced, i.e., twice as many EU students compared to US students, and the sample size was low, we did not include statistical analyses to assess whether the combination of two independent variables had an effect on a dependent variable, i.e., accuracy or time needed. Therefore, multiple one-way ANOVA comparison of mean were or $\chi^2$-test when comparing accuracy were performed (Table 1). The results show no statistical difference between G1, G2, G3 groups in time needed to read abstracts and answer questions for US students independent of UC1 or UC2.

Looking at EU students, a statistically significant difference in time needed was observed in UC1 (F=13.466, p<0.001) as well as UC2 (F=9.705, p<0.001), where post hoc pairwise comparisons using Tukey HSD showed that for both UC1 and UC2 the G1 group needs more time than G3 and in the case of UC2 the G1 group needs longer than G2.

Comparing accuracy (percentages) with $\chi^2$-test showed that for EU students, there were statistically significant differences in both UC1 ($\chi^2$=48.349, p<0.001) as well as UC2 ($\chi^2$=23.14, p<0.001), where a post hoc pairwise comparisons showed that for both UC1 and UC2 the accuracy is higher in G1 compared to either G2 or G3.

Similarly, for US Students, the results showed there were statistically significant differences in both UC1 ($\chi^2$=6.778, p=0.034) as well as UC2 ($\chi^2$=12.905, p=0.002), where the post hoc pairwise comparisons in UC2 showed that accuracy in G1 is higher than in G3, but for UC1 we observed that G2 is higher than G3.

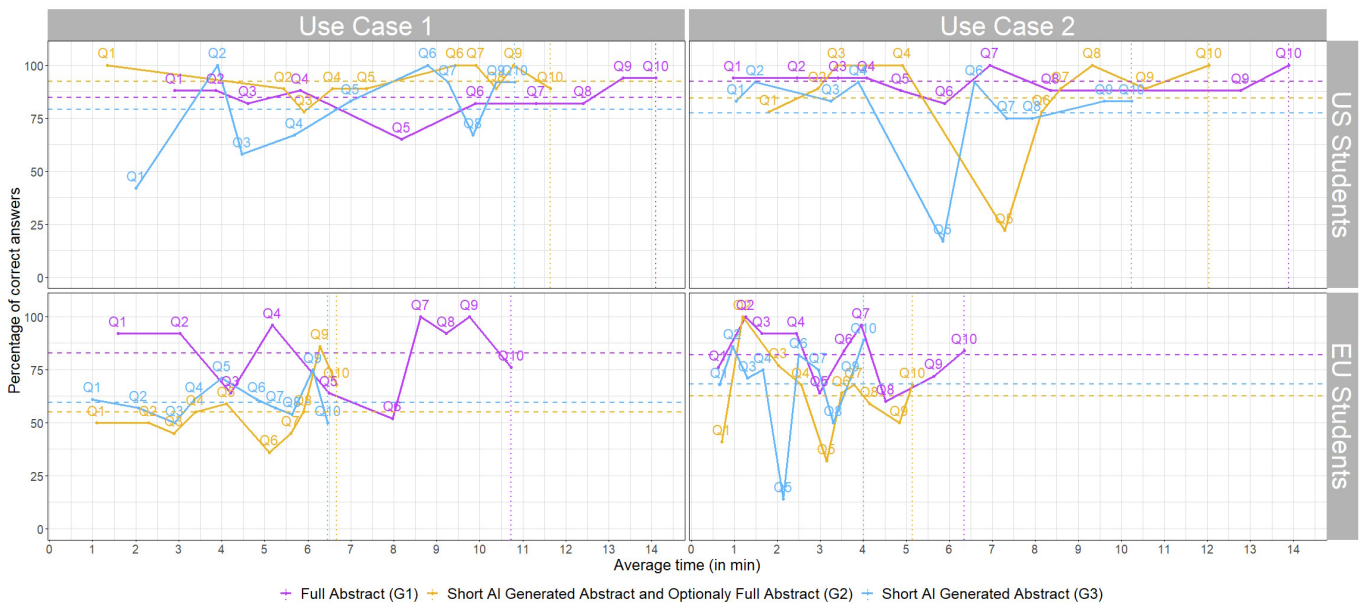

**Figure 5: Comparison of US and EU students in accuracy (y-axis) and time (length of line arcs on the x-axis) needed to screen the scientific abstracts and extract relevant information.**

| Use Case | Time [sec] | | | | | Accuracy [proportion] | | | | |
|---|---|---|---|---|---|---|---|---|---|---|
| | G1 | G2 | G3 | One-way Anova | Turkey HSD* | G1 | G2 | G3 | Chi-square | Pairwise comaprison* |
| **EU Students** | | | | | | | | | | |
| Use case 1 | 643.3 [557.81, 728.8] | 400.63 [316.79, 484.47] | 387.74 [312.85, 462.64] | 13.466 (p<0.001) | G1>G2, G1>G3 | 0.828 [0.775, 0.873] | 0.55 [0.482, 0.617] | 0.596 [0.536, 0.654] | 48.349 (p<0.001) | G1>G2, G1>G3 |
| Use case 2 | 380.95 [328.46, 433.44] | 308.93 [269.38, 348.47] | 240.29 [192.25, 288.33] | 9.705 (p<0.001) | G1>G3 | 0.82 [0.767, 0.866] | 0.627 [0.56, 0.691] | 0.682 [0.624, 0.736] | 23.14 (p<0.001) | G1>G2, G1>G3 |
| **US Students** | | | | | | | | | | |
| Use case 1 | 845.99 [358.24, 1333.75] | 698.26 [263.09, 1133.42] | 649.02 [418.04, 879.99] | 0.288 (p=0.752) | | 0.847 [0.784, 0.898] | 0.922 [0.846, 0.968] | 0.792 [0.708, 0.86] | 6.778 (p=0.034) | G2>G3 |
| Use case 2 | 833.41 [253.48, 1413.34] | 722.72 [333.86, 1111.59] | 614.4 [224.31, 1004.49] | 0.224 (p=0.8) | | 0.924 [0.873, 0.959] | 0.844 [0.753, 0.912] | 0.775 [0.69, 0.846] | 12.905 (p=0.002) | G1>G3 |

*Bonferroni correction for multiple comparisons

**Table 1: Quantitative results for all ten questions and both use case.**

## 4 Discussion and Conclusions

Lifestyle-related diseases such as cardiovascular diseases, type 2 diabetes, or cancer have a major impact on quality of life and are the major causes of years of life lost worldwide [6]. Due to the heavy workload of family care team members, complications that could often be mitigated by targeted lifestyle modification advice from the family health team are often not provided to patients. Research suggests that controlling these modifiable risk factors could prevent up to 80% of premature heart disease, stroke and diabetes [7]. As the era of big data per individual emerges, healthcare professionals will no longer be able to process the rapidly growing information on new developments and knowledge in their field [8]. Health system improvement in primary care is typically slow, in part due to the reliance on passive knowledge dissemination and the absence of a systematic approach to identify the gaps between evidence and practice and to implement appropriate interventions to close these gaps [9,10]. Continuously integrating innovations and evidence-based practices (EBPs) into routine care settings is the core of Learning Health Systems (LHSs). In an LHS, the EBPs should integrate the evidence synthesized from existing scientific literature and generated from the data obtained inside the institution's data networks [11].

Abstract Text Summarization (ATS) is one of the potential downstream tasks, driven by the massive increase in the availability of textual documents such as blogs, news articles and reports [12]. In the health sector, ATS has typically consisted of models attempting to extract strong evidence from large datasets of biomedical data to support complex clinical decisions, where many such models suffered from a lack of ability to capture the clinical context, quality of evidence, and purposeful selection of text for summarization [13].

There are well-known issues with the accuracy of the output and the tendency of some models to 'hallucinate', despite the fact that output quality is typically one of the goals of current state-of-the-art models [14,15].

Our study shows that using generative AI for screening novel literature is effective and efficient and can be used to save time. On the other hand, we also show that a significant drop in accuracy related to extracted knowledge was present in most cases. It is important to acknowledge that this report represents the work in progress and we plan to continue our research to further refine our findings. LLMs could significantly contribute to reducing the time needed to keep up to date with the recent scientific literature for healthcare providers. However, it should be noted that a significant effort in terms of human-based evaluation of the results from generative AI will be needed in the coming years to improve trust in generative AI among healthcare professionals. Looking even further into the future, generative AI will play an important role in extracting data from novel scientific literature to feed it into the existing data networks used to enable data-driven learning health systems. Again, the question of how to maintain a high quality of extracted data remains an open question. Additional research will be conducted in the future to examine additional use cases beyond those presented in UC1 and UC2 and to make additional comparisons based on the difficulty of the questions.

## ACKNOWLEDGMENTS
The authors (GS, PK, LG) acknowledge partial support from the Slovenian Research Agency (ARRS N3-0307, ARRS P2-0057 and ARRS BI-US/22-24-138).

## REFERENCES
[1] Lenfant C. Shattuck lecture–clinical research to clinical practice–lost in translation? N Engl J Med. 2003;349(9):868–74.
[2] Davis D, Evans M, Jadad A, Perrier L, Rath D, Ryan D, et al. The case for knowledge translation: shortening the journey from evidence to effect. BMJ. 2003;327(7405):33–5.
[3] Kilbourne AM, Goodrich DE, Miake-Lye I, Braganza MZ, Bowersox NW. Quality enhancement research initiative implementation roadmap: toward sustainability of evidence-based practices in a learning health system. Medical care. 2019 Oct;57(10 Suppl 3):S286.
[4] Cachola, I., Lo, K., Cohan, A., & Weld, D. S. (2020). TLDR: Extreme summarization of scientific documents. arXiv preprint arXiv:2004.15011.
[5] Kocbek P, Gosak L, Musović K, Stiglic G. Generating Extremely Short Summaries from the Scientific Literature to Support Decisions in Primary Healthcare: A Human Evaluation Study. Artificial Intelligence in Medicine: 20th International Conference on Artificial Intelligence in Medicine, AIME 2022, Halifax, NS, Canada, June 14–17, 2022, Proceedings 2022 (Jun 14), 373-382.
[6] Li Y, Schoufour J, Wang DD, Dhana K, Pan A, Liu X, Song M, Liu G, Shin HJ, Sun Q, an Al-Shaar L, 2020. Healthy lifestyle and life expectancy free of cancer, cardiovascular disease, and type 2 diabetes: prospective cohort study. BMJ 368 (Jan 2020), 6669. DOI: 10.1136/bmj.l6669
[7] White ND, Lenz TL, and Smith K. Tool guide for lifestyle behavior change in a cardiovascular risk reduction program. Psychology research and behavior management 19, 6 (Aug 2013), 55-63. DOI: 10.2147/PRBM.S40490

[8] Steinhubl SR, and Topol EJ. Moving from digitalization to digitization in cardiovascular care: why is it important, and what could it mean for patients and providers?. Journal of the American College of Cardiology 66, 13 (Sep 2015), 1489-96, DOI: 10.1016/j.jacc.2015.08.006

[9] Lenfant C, 2003. Shattuck lecture–clinical research to clinical practice–lost in translation? N Engl J Med 349, 9 (Aug 2003), 868–74, DOI: 10.1056/NEJMsa035507

[10] Davis D, Davis ME, Jadad A, Perrier L, Rath D, Ryan D, Sibbald G, Straus S, Rappolt S, Wowk M, and Zwarenstein M. The case for knowledge translation: shortening the journey from evidence to effect. BMJ 327, 7405 (Jul 2003), 33–5. DOI: 10.1136/bmj.327.7405.33

[11] Kilbourne AM, Goodrich DE, Miake-Lye I, Braganza MZ, and Bowersox NW, 2019. Quality enhancement research initiative implementation roadmap: toward sustainability of evidence-based practices in a learning health system. Medical care 57, Suppl 3 (Oct 2019), S286, DOI: 10.1097/MLR.0000000000001144

[12] Shi T, Keneshloo Y, Ramakrishnan N, and Reddy CK, 2021. Neural abstractive text summarization with sequence-to-sequence models. ACM Transactions on Data Science 2, 1, 1-37. DOI: https://doi.org/10.1145/3419106

[13] Afzal M, Alam F, Malik KM, and Malik GM, 2020. Clinical context–aware biomedical text summarization using deep neural network: model development and validation. Journal of medical Internet research 22, 10 (Oct 2020), e19810. DOI: 10.2196/19810

[14] Marcheggiani D, Perez-Beltrachini L, 2018. Deep Graph Convolutional Encoders for Structured Data to Text Generation. Proceedings of the 11th International Conference on Natural Language Generation. DOI: 10.18653/v1/W18-6501

[15] Rohrbach A, Hendricks LA, Burns K, Darrell T, AND Saenko K, 2018. Object hallucination in image captioning. Proceedings of the 2018 Conference on Empirical Methods in Natural Language Processing. DOI: 10.18653/v1/D18-1437